\documentclass[10pt,twocolumn,letterpaper]{article}

\usepackage[pagenumbers]{wacv} 

\usepackage{graphicx}
\usepackage{amsmath}
\usepackage{amssymb}
\usepackage{booktabs}
 
\usepackage{listings}

\usepackage[pagebackref,breaklinks,colorlinks]{hyperref}

\usepackage[capitalize]{cleveref}
\crefname{section}{Sec.}{Secs.}
\Crefname{section}{Section}{Sections}
\Crefname{table}{Table}{Tables}
\crefname{table}{Tab.}{Tabs.}

\def\wacvPaperID{1620} 
\def\confName{WACV}
\def\confYear{2025}

\usepackage[acronym,nomain]{glossaries}
\makeglossaries
 
 \newacronym{cnn}{CNN}{Convolutional Neural Networks}
 \newacronym{wsss}{WSSS}{Weakly Supervised Semantic Segmentation}
\newacronym{wsvos}{WSVOS}{Weakly Supervised Video Object Segmentation}
\newacronym{gwsvos}{GWSVOS}{Generalized Weakly Supervised Object Segmentation}
\newacronym{htopv}{HTOPV}{Highly Transient Object Presence  Video}
\newacronym{copv}{COPV}{Consistent Object Presence Video}
\newacronym{tcn}{TCN}{temporal convolutional network}
\newacronym{top}{TOP}{Transient Object Presence}
\newacronym{cop}{COP}{Consistent Object Presence}

\newacronym{wsvostop}{WSVOS-TOP}{Transient Object Presence}
\newacronym{wsvoscop}{WSVOS-COP}{Consistent Object Presence}

\newacronym{cam}{CAM}{Class Activation Map}
\newacronym{sam}{SAM}{Segment Anything Model}
\newacronym{stcam}{ST-CAM}{spatio-temporal class activation map}
\newacronym{vstd}{VDST-Net}{video spatio-temporal disentanglement networks}
\newacronym{tcam}{TCAM}{Temporal class activation map}
\newacronym{vit}{ViT}{Vision Transformer}
\newacronym{mlp}{MLP}{Multilayer Perceptron}
\newacronym{mae}{MAE}{Masked Auto-Encoder}
\newacronym{kd}{KD}{Knowledge Distillation}
\newacronym{dcrf}{DCRF}{Dense Conditional Random Field}
\newacronym{iou}{IoU}{ Intersection over Union }

\usepackage{amsmath,amssymb,amsfonts}
\usepackage{graphicx}
\usepackage{textcomp}
\def\BibTeX{{\rm B\kern-.05em{\sc i\kern-.025em b}\kern-.08em
    T\kern-.1667em\lower.7ex\hbox{E}\kern-.125emX}}

\usepackage{xargs}                      
\usepackage[pdftex,dvipsnames]{xcolor}  
\usepackage[colorinlistoftodos,prependcaption,textsize=small]{todonotes}
\newcommandx{\unsure}[2][1=]{\todo[linecolor=red,backgroundcolor=red!25,bordercolor=red,#1]{#2}}
\newcommandx{\change}[2][1=]{\todo[linecolor=blue,backgroundcolor=blue!25,bordercolor=blue,#1]{#2}}
\newcommandx{\info}[2][1=]{\todo[linecolor=OliveGreen,backgroundcolor=OliveGreen!25,bordercolor=OliveGreen,#1]{#2}}
\newcommandx{\improvement}[2][1=]{\todo[linecolor=Plum,backgroundcolor=Plum!25,bordercolor=Plum,#1]{#2}}
\newcommandx{\thiswillnotshow}[2][1=]{\todo[disable,#1]{#2}}

\usepackage{hyperref}
\hypersetup{
    colorlinks=true,
    linkcolor=black, 
    filecolor=blue,
    urlcolor=blue,
    citecolor=green,
}
 
\usepackage{multirow}
\usepackage{tabularx}
\usepackage{hhline}
\usepackage{array}
\newcolumntype{P}[1]{>{\centering\arraybackslash}p{#1}}
\newcolumntype{L}[1]{>{\raggedright\let\newline\\\arraybackslash\hspace{0pt}}m{#1}}
\newcolumntype{C}[1]{>{\centering\let\newline\\\arraybackslash\hspace{0pt}}m{#1}}
\newcolumntype{R}[1]{>{\raggedleft\let\newline\\\arraybackslash\hspace{0pt}}m{#1}}
\usepackage{booktabs}

\usepackage{paralist}

\usepackage[normalem]{ulem}
\usepackage{xfrac}

\usepackage[symbol]{footmisc}

\RequirePackage[normalem]{ulem} 
\RequirePackage{xcolor}
\usepackage{soul}
\usepackage{ulem}
\usepackage{censor}
\newlength\nextcharwidth
\makeatletter
\renewcommand\@cenword[1]{%
  \setlength{\nextcharwidth}{\widthof{#1}}%
  \censorrule{\nextcharwidth}%
  \kern -\nextcharwidth%
  #1}
\makeatother

\providecommand{\DIFadd}[1]{{{{\protect{#1}}}}} 
\providecommand{\DIFdel}[1]{{}}    

\usepackage{algorithmicx}
\usepackage[ruled]{algorithm}
\usepackage{algpseudocode}

\graphicspath{{figures/}}
\DeclareGraphicsExtensions{.pdf,.jpeg,.png,.eps}

\RequirePackage{suffix}
\long\def\RC#1\par{\makebox[0pt][r]{\bf  \hspace{4mm}}\textbf{\textit{#1}}\par} 
\WithSuffix\long\def\RC*#1\par{\textbf{\textit{#1}}\par} 
\long\def\AR#1\par{\makebox[0pt][r]{ \hspace{10pt}}#1\par} 
\WithSuffix\long\def\AR*#1\par{#1\par} 

\begin{document}

\title{Disentangling spatio-temporal knowledge for weakly supervised object detection and segmentation in surgical video}


\author{
Guiqiu Liao\textsuperscript{1} \and
Matjaz Jogan\textsuperscript{1}  \and
Sai Koushik\textsuperscript{1,3}  \and
Eric Eaton\textsuperscript{2} \and
Daniel A. Hashimoto\textsuperscript{1,2}
\\
\textsuperscript{1}Penn Computer Assisted Surgery and Outcomes Laboratory, \\ Department of Surgery, University of Pennsylvania\\
\textsuperscript{2}Department of Computer and Information Science, University of Pennsylvania\\
\textsuperscript{3}Department of Electrical and Systems Engineering, University of Pennsylvania
}

%
%
 
\maketitle

\begin{abstract}
 
 Weakly supervised video object segmentation (WSVOS) enables the identification of segmentation maps without requiring \DIFadd{extensive annotations} of object masks, relying instead on coarse video labels indicating object presence. \DIFadd{Weakly supervised semantic segmentation of objects in surgical videos is however more challenging due to a complex interaction of multiple transient objects, such as surgical tools moving in and out of the surgical field. In this scenario, state-of-the-art WSVOS methods struggle to learn accurate segmentation maps. We address this problem by introducing Video Spatio-Temporal Disentanglement
    Networks (VDST-Net), a framework to disentangle complex spatiotemporal object interactions using semi-decoupled knowledge distillation to predict high-quality class activation maps (CAMs).}  \DIFadd{A teacher network is designed to help a temporal-reasoning student network resolve activation conflicts, as the student leverages temporal dependencies when specifics about object location and timing in the video are not provided}. We demonstrate the efficacy of our framework on a challenging surgical video dataset where objects are, on average, present in less than 60\% of annotated frames, and compare our method to state-of-the-art methods on surgical data and on a public dataset commonly used to benchmark WSVOS. Our method outperforms state-of-the-art techniques and generates accurate segmentation masks under video-level weak supervision. 
\end{abstract}


\section{Introduction} 
\label{sec:intro}

Incorporating expert knowledge into machine learning for surgical video analysis demands extensive and costly annotations. Annotated surgical video databases are rare but crucial for applications ranging from video review and analysis to AI supported robotic assistance~\cite{volkov2017machine}.

The scarcity of annotated video is not unique to surgery. To address this,~\gls{wsss} methods were developed~\cite{zhou2016learning,kumar2017hide,gao2021ts,kweon2024sam}, using image-level class labels to identify areas in an image that contain the corresponding object. Such methods have also been applied to video with frame-level labels, exploiting temporal information to improve accuracy~\cite{nwoye2019weakly,liu2021weakly,yudistira2022weakly}.

\begin{figure*}[!t]
\includegraphics[width=0.95\textwidth]{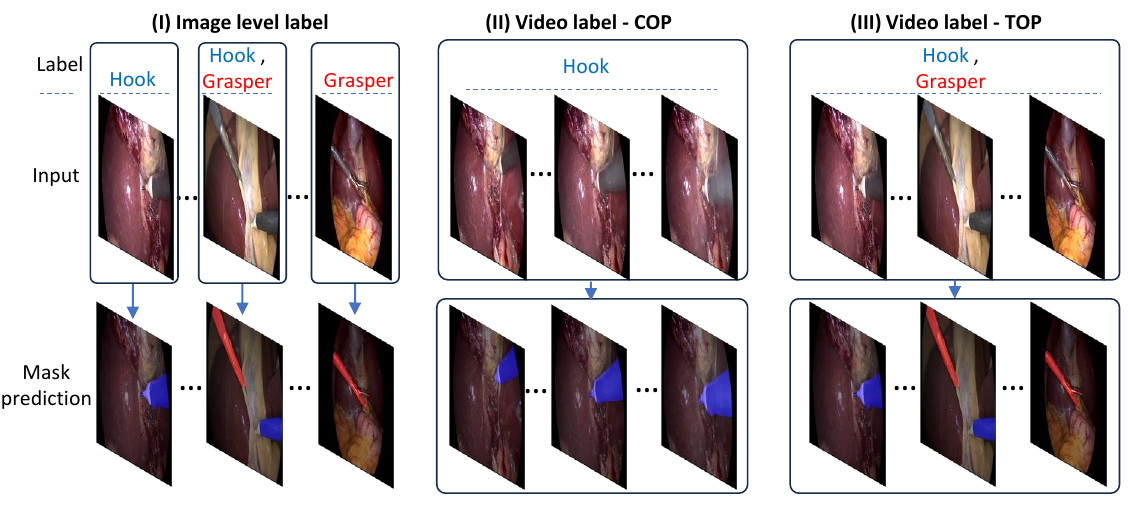}
\centering
\caption{Three types of weakly supervised detection/segmentation in video. Type I has image-level presence labels available, and research for this task usually builds on WSSS adding temporal  constraints to improve prediction. Type II \gls{wsvos}-\gls{cop} only has video-level labels, and the object is assumed to be in the video for most of the frames (this is the scenario for YouTube-object dataset). Type III \gls{wsvos}-\gls{top} is the most challenging, where the label only indicates the presence of objects for the whole video, yet each object may be present in the video temporarily (i.e., in only a subset of frames).
} 
\label{fig_task}
\end{figure*}

In \DIFadd{the more difficult problem of} \gls{wsvos}, object \DIFadd{presence} labels are only available at video level, \DIFadd{while no image level presence or segmentation supervision signal is provided.}\DIFdel{introducing,} \DIFadd{This introduces higher} uncertainty about object presence in each frame. Some approaches to WSVOS leverage motion cues ~\cite{tokmakov2016weakly, zhang2020spftn}. However, these methods often lack robustness against background motion and stationary objects. A recent end-to-end  framework~\cite{belharbi2023tcam} has achieved state-of-the-art performance in object detection on two challenging datasets. In these datasets, the primary objects or concepts present in the video are labeled and are constantly present in most frames. We refer to this type of problem as \gls{cop}, in contrast to \gls{top} where no assumptions are made about object constancy (Figure \ref{fig_task}). The latter case covers the significant variation in temporal presence of objects in surgical videos, where  annotations are often based on the start and end times of surgical steps with annotated tools intermittently moving in and out of view. In \gls{wsvostop}, the algorithm must thus learn only from object classes without any implicit or explicit information on location, size, or temporal presence, and without assumptions about a specific count of object classes per frame.

\DIFadd{In order to address these challenges,}\DIFdel{. In this work,} we propose a \DIFadd{novel} method for \gls{wsvos} \DIFdel{that addresses ...\gls{cop} and \gls{top} problems.} \DIFadd{that excels at the \gls{top} problem on surgical data while also achieving state-of-the-art performance in \gls{cop} scenarios.}
Our contributions include:
\begin{enumerate}
   \item We propose an efficient network architecture to generate high-quality spatio-temporal  class activation maps trained with video-level classification labels.
    \item We introduce \gls{vstd}, which captures information associations across time and effectively addresses activation conflicts due to the complex interaction of multiple target objects, including objects appearing and disappearing from the video.

\item Experiments on our \gls{top} surgical video dataset based on  Cholec80~\cite{twinanda2016endonet} with an average object presence rate of under 60\% demonstrate a clear advantage of our method in \gls{wsvos}-\gls{top}. 

 \item \DIFadd{Experiments on YouTube-Objects
(v2.2) demonstrate that our method also outperforms state-of-the-art approaches  in classic \gls{wsvos}-\gls{cop}.}
\end{enumerate}

\section{Related Work} 
 
\subsection{Weakly supervised segmentation for image data}
Image-level weakly supervised semantic segmentation has been well-researched, and most existing methods are based on \gls{cam} \cite{zhou2016learning}. Raw CAM may be sufficient for detection tasks, but it fails to delineate the target object at the semantic level as it tends to learn the most salient feature. Training strategies such as randomly masking out images\cite{kumar2017hide} or feature maps\cite{hou2018self,lee2019ficklenet} can refine the CAM by enhancing activation of novel features. Loss functions designed to improve CAM have been proposed, including SEC loss \cite{kolesnikov2016seed}, CRF loss \cite{tang2018regularized}, regularization
losses \cite{yao2021non}, and contrastive loss \cite{yang2024separate}. Recently, works on network modeling that explore the long-range dependence and attention mechanism of \gls{vit} in CAM \cite{gao2021ts,xu2022multi} demonstrated \gls{vit} based \gls{cam} could outperform \gls{cnn} based models in image semantic segmentation tasks. There are emerging works that integrate foundation models with weakly supervised segmentation. \cite{lin2023clip} uses the Contrastive Language-Image Pretraining (CLIP) \cite{radford2021learning} for \gls{wsss}. 
\gls{sam} was also applied to WSSS \cite{kweon2024sam}. While these methods improve segmentation performance for tasks with image-level labels, they are not directly applicable to weakly supervised segmentation tasks in video. 

\subsection{Weakly supervised detection and segmentation for video}
\gls{wsvos} can be categorized into three types of tasks, as shown in Figure \ref{fig_task}. In the first type (Type I), frame-level object presence labels are provided. Research in this area often incorporates motion constraints to enhance prediction accuracy, as demonstrated in studies such as \cite{liu2021weakly, li2019motion,yudistira2022weakly}.  \cite{nwoye2019weakly} employs LSTM to reason across time for detection and tracking in surgical videos.\cite{hou2017tube} adapts a 3D CNN to incorporate temporal dependencies for detection in videos.
The second type of \gls{wsvos} (Type II) has object labels available only at the video-level, thus not explicitly determining object presence in a particular frame. \DIFadd{However, a common assumption in the literature in the WSVOS-COP setting is that labeled objects are present throughout the entire video, with frames missing these objects treated as noise. This scenario is exemplified by well-known datasets such as YouTube-Objects}\cite{kalogeiton2016analysing}.Consequently, this setup is close to a Type I problem as frame content can be inferred from video-level annotations. Recent work on TCAM \cite{belharbi2023tcam} extracts the most reliable pseudo-semantic supervision signal in video to train a subsequent network for detection in individual images, efficiently addressing object detection for this case, while still showing limited robustness to smaller objects and temporal inconsistencies.
In the third type (Type III), labels merely indicate the general presence of objects throughout an entire video, despite each object possibly appearing only temporarily. This is depicted in the right column of Figure \ref{fig_task}. Consequently, the presence label assigned to a video may not accurately reflect a large subset of frames, introducing \DIFadd{higher uncertainty that leads to} challenge in learning generalized representations. Using motion cues could not resolve such an issue when including scenarios where the camera can occasionally move while the objects stay still. \DIFadd{Existing semantic segmentation methods that use sparse or coarse labels  to segment surgical video only focus on the \gls{wsss} problem with presence labels for individual frames available}\cite{nwoye2019weakly,yang2022weakly,lee2019weakly}. \DIFadd{Zero-shot approaches rely on \gls{sam} and a prompting strategy}\cite{yue2024surgicalsam}. \DIFadd{However, no attempts have been made to apply WSVOS to surgical videos or similar TOP scenarios using a single class presence label per video. }  To the best of our knowledge, our work is the first to address this type of \gls{top} data and train an end-to-end network to predict segmentation masks for the video as a whole.

\subsection{Network reasoning across time and knowledge distillation}
3D CNN \cite{tran2015learning}, recurrent neural networks (RNN) \cite{sherstinsky2020fundamentals}, and temporal Transformers \cite{yan2021learning} are widely used architectures that capture time dependency and have been applied to various video and volumetric imaging tasks \cite{li2019multi, zhu2021weakly, liu2022video}. However, due to the lack of direct supervision signals, learning spatiotemporal features for weakly supervised segmentation is difficult.  \gls{kd} \cite{hinton2015distilling} using a teacher-student network was originally designed for model compression and can gradually improve network learning capacity \cite{furlanello2018born}. This property allows KD to enhance learning in weakly or semi-supervised tasks. L2G \cite{jiang2022l2g} applied KD to extend CAM from local to global. SCD \cite{xu2023self} uses KD to incorporate feature correspondence for CAM refinement in images. Seco \cite{yang2024separate} employs KD to decouple occurrences for refining CAM in images. Our method employs KD to address the challenges \DIFadd{of} ill-posed video-level weak supervision, which causes activation conflicts of features that are not associated with the correct class. 

We use a constrained teacher to allow the network not only to produce correct activation, but also to refine the backbone to learn generalized features for new domain-specific data.  Decoupled KD \cite{zhao2022decoupled} has proven to be a versatile method for flexibly distilling meaningful knowledge. Our design, which could be categorized as being based on semi-decoupled KD, is tailored for video weakly supervised segmentation. Rather than relying on soft labels from the teacher, we enable the teacher's predictions while utilizing ground truth to form gates for student supervision. This approach allows the student to learn effectively from an imperfect teacher while still gaining temporal reasoning abilities.

\begin{figure*}[h!]
\includegraphics[width=1.0\textwidth]{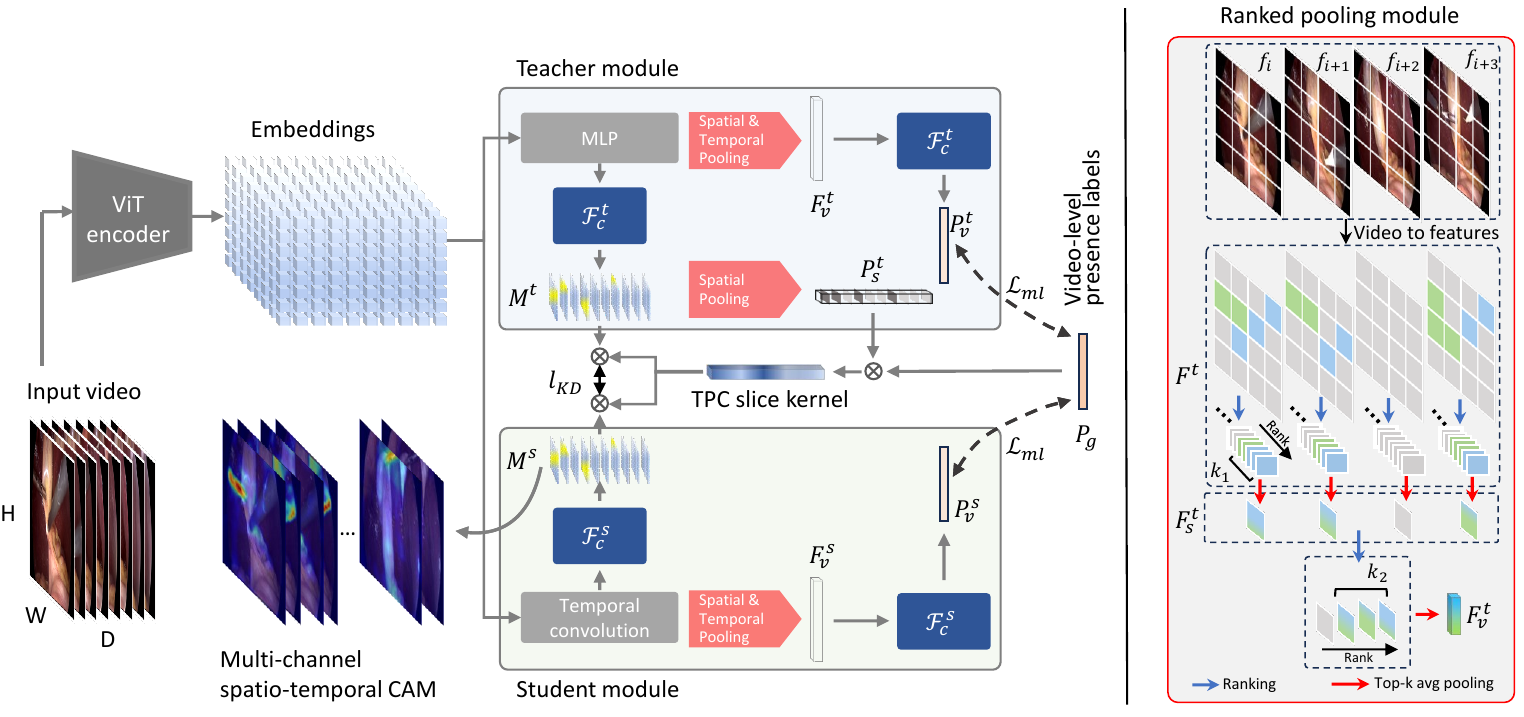}
\centering
\caption{Left: Our \gls{vstd} approach deploying knowledge distillation to disentangle spatial and temporal information for weakly supervised learning with video-level labels. \DIFadd{Ground truth video presence labels $P_g$ provide supervision to both teacher and student, while knowledge in activation map $M^t$ is transferred from teacher to student. Right: Ranked spatial and temporal pooling captures information about multiple objects in a frame while filtering out spurious information from frames where a target object is missing.}
} 
\label{fig_m}
\end{figure*}
  
\section{Methods} 
\label{sec:methods}

Figure \ref{fig_m} (left) illustrates the proposed \gls{vstd} framework. 
The core of the framework consists of a teacher-student network pair designed to disentangle spatial and temporal knowledge. Both modules utilize a \gls{stcam} and share input from \gls{vit}~\cite{dosovitskiy2020image} \DIFadd{encoded frames and video-level class labels as the supervision signal}. The primary distinction between the teacher and the student lies in \DIFadd{their upper-layer feature extraction modules: the teacher employs a \gls{mlp}, while the student employs a \gls{tcn}. This design intentionally restricts the teacher from temporal interactions, reducing the potential for activation conflicts between frames, while progressively increasing the student's ability to reason across time.}  
\DIFadd{We utilize a semi-decoupled online distillation approach, where both teacher and student modules are trained using the same video labels, and efficient knowledge transfer is facilitated through gated activation maps. While the teacher module has a powerful backbone capable of generating highly accurate activation maps, it still exhibits detection gaps and cannot consistently provide fully reliable knowledge for distillation. To resolve these limitations, the semi-decoupled distillation allows the student module to selectively learn from the teacher's gated outputs under the constraints of video-level ground truth, while enhancing temporal consistency.}
 

\subsection{Video encoder}
A \gls{vit} encoder is used to encode patches from individual frames of the input video $V \in {\mathbb{R}}^{W\times H \times D}$, producing a stack of token embeddings reshaped as \DIFdel{$T$}$X \in {\mathbb{R}}^{w\times h \times D \times C_e}$, where $w=W/P$, $h=H/P$, and $P$ denotes the width/height of a patch, and $C_e$ is the embedding dimension.
 \DIFdel{$T$}\DIFadd{$X$} captures features with strict temporal correspondence (maintaining the temporal dimension $D$) and high spatial correspondence. 

\subsection{Teacher module}
\DIFadd{As depicted in the blue block of Fig \ref{fig_m},}\DIFdel{$T$}\DIFadd{video embedding tensor $X$} is further processed by a \gls{mlp} block that linearly projects \DIFdel{$T$}\DIFadd{$X$} \DIFdel{to $C$ channels of} a feature map $F^t \in {\mathbb{R}}^{w\times h \times D \times C}$ \DIFadd{with $C$ channels }, maintaining spatio-temporal correspondence. Following a line of work on ViT-based 2D \gls{cam} \cite{gao2021ts,xu2022multi}, spatial pooling is applied to $F^t$ to obtain a slice-wise feature map $F_s^t \in {\mathbb{R}}^{1\times 1 \times D \times C}$. Then, temporal pooling is applied to acquire a final feature map for video classification $F_v^t \in {\mathbb{R}}^{1\times 1 \times 1 \times C}$. Inspired by weakly supervised temporal scene anomaly localization~\cite{tian2021weakly} we use a ranked top-$k$ average pooling\DIFadd{:}
\begin{equation} 
\DIFadd{ F_{s}^{t,i}=  \frac{1}{k_1} \sum\nolimits_{j=1}^{k_1}{F^{t,r_j^i}}\enspace }
\label{eq_L_s}
\end{equation} 
\DIFadd{where $i \in [1,D]$ is the temporal index of feature $F_{s}^{t}$. $r_j^i\in [1,H\times W]$ is the ranked spatial index of $F^{t}$ at a given temporal slice.
This approach first ranks all the spatial patch features by magnitude, and then selects the top-$k_1$ patches for average pooling. 
Following a similar strategy, ranked temporal pooling is applied as:}
\begin{equation} 
\DIFadd{ F_{v}^{t}=  \frac{1}{k_2} \sum\nolimits_{l=1}^{k_2}{F_s^{t,r_l}}\enspace }
\label{eq_L_s}
\end{equation} 
\DIFadd{where $r_l\in [1,D]$ is the ranked temporal index of $F_s^{t}$.
$k_2$ is used to select top temporal features. The ranked spatio-temporal pooling process is illustrated in the Fig. \ref{fig_m} (right).}
\DIFadd{Maximum pooling or global average pooling can be seen as extreme cases of ranked pooling (i.e. $k_2$=1 or $k_2= D$), which encourage activating either on the most salient feature location or whole video, causing low accuracy in the surgical video \gls{top} scenario.   
Ranked pooling design can mitigate such issues and force the ST-CAM to activate on the proper temporal slices and area of interest. }
Finally, with a fully connected classification layer $\mathcal{F_C}^t$, the final video prediction $P_v^t \in {\mathbb{R}}^{1\times 1 \times 1 \times N}$ is obtained by $P_v^t = \mathcal{F_C}^t(F_v^t)$, where $N$ is the number of classes. Plugging $\mathcal{F_C}^t$ channel-wise to feature map $F^t$ results in a  multi-channel \gls{stcam} $ M^t \in {\mathbb{R}}^{w\times h \times D \times N}$. \DIFadd{To help the backbone learn generalized features at different scales, after each \gls{mlp} layer, features are randomly downsampled during training with a probability of 0.5.}

\subsection{Student module} 
Employing an \gls{mlp} for encoding of embeddings \DIFadd{$X$}\DIFdel{$T$}, \DIFadd{the teacher module} processes each frame in isolation, thus eliminating temporal conflicts. However, \DIFadd{this}\DIFdel{it} fails to exploit \DIFadd{the} temporal coherence to enhance prediction robustness. \DIFadd{We thus designate a temporal convolution-based network in the student module. As depicted in Fig. \ref{fig_m}, the student network adopts the same spatio-temporal pooling mechanism from the teacher and trains an additional fully connected classification layer $\mathcal{F_C}^s$. This layer produces the student's video prediction $P_v^s \in {\mathbb{R}}^{1\times 1 \times 1 \times N}$ and the student \gls{stcam} $M^s \in {\mathbb{R}}^{w\times h \times D \times N}$. The student's} \gls{tcn} can infer associations between frames, yet will introduce errors due to the absence of frame-level annotation. To mitigate this issue, we utilize a semi-decoupled \gls{kd} technique that refines a student network with higher capacity through efficiently selected supervision signals.

\subsection{Training loss and semi-decoupled knowledge distillation} 

 \DIFadd{The teacher module optimizes on a multi-label soft margin classification loss $\mathcal{L}_{ml}(P_v^t, P_g)$ utilizing ground truth video labels $P_g \in \mathbb{R}^{1 \times N}$. The student module optimizes the same classification loss and a KD loss that further refines the student's activation map.}
 \DIFdel{To help...} We employ a semi-decoupled \cite{zhao2022decoupled} approach to knowledge distillation where weak video ground truth labels supervise both teacher and student networks' predictions while applying KD through activation maps. \DIFadd{\gls{kd} was used to transfer teacher's prediction logits as soft labels}\cite{hinton2015distilling,furlanello2018born}, \DIFadd{or attention}\cite{zagoruyko2016paying,guo2023class} \DIFadd{; our approach is closer to latter, as it uses attention maps to guide the student}\cite{guo2023class}.\DIFdel{... map.} \DIFadd{KD} loss is computed by considering only the true positive channel (TPC) and positive slice predictions from the teacher network.
 
 Initially, a teacher slice prediction $P_s^t \in {\mathbb{R}}^{1\times 1 \times D \times N}$ is obtained by applying maximum spatial pooling to the teacher \gls{stcam}. Subsequently, we create a TPC slice kernel $K_{tp} \in {\mathbb{R}}^{1\times 1 \times D \times N}$ through the operation $K_{tp}= P_s^t \otimes r(P_g)$, where $r(\cdot)$ represents a broadcast operation to match dimensions for element-wise multiplication $\otimes$.
 A final gated \gls{kd} loss is calculated by:
\begin{equation} 
 {l}_{KD} = \mathcal{L}_{mse}(M^s \otimes r(K_{tp}), M^t \otimes r(K_{tp})) \enspace 
\label{eq_L_kd}
\end{equation} 
where $\mathcal{L}_{mse}$ is a Mean Squared Error loss function, and $r(\cdot)$ here matches the dimension of $K_{tp}$ to $M^t$ and $M^s$. In analogy, this operation focusing exclusively on frames where true positive pseudo masks are predicted, akin to singling out specific instances for loss calculation.
The final loss used for optimizing the student module is given by:
\begin{equation} 
 {l}_{s} = \mathcal{L}_{ml}(P_v^s, P_g) + \alpha {l}_{KD}  \enspace 
\label{eq_L_s}
\end{equation} 
In Eq. (\ref{eq_L_s}) $\alpha$ is a scaling factor for the distillation loss. Finally, we apply a ReLU layer and normalization to the \glspl{stcam} extracted from classification networks.   

\section{Experiments} 
\label{sec:experiments_results}

\subsection{Datasets}
To investigate the characteristics of our proposed method, we utilize two distinct video dataset collections: \DIFadd{ Cholec80}\cite{twinanda2016endonet} \DIFadd{dataset and YouTube-Objects video dataset (v2.2)} \cite{kalogeiton2016analysing}. The Cholec80 dataset consists of 80 cholecystectomy surgery videos, each downsampled to 1 FPS, with presence labels of seven surgical tools ($N=7$) in each frame. We create a variant \gls{top} video dataset based on Cholec80.  First these videos are split into 30-second clips without overlap, merging the tool labels across frames to generate a video-level tool presence binary vector label for each clip. The test set is based on CholecSeg8K \cite{hong2020cholecseg8k} which contains 8K images from 17 videos in Cholec80 with segmentation labels for instruments and anatomy. From this dataset we derive 100 video clips containing 3K frames. To prevent contamination between testing and training data, 5,296 training clips are sampled from from 63 Cholec80 videos not included in CholecSeg8k. We calculate the tool frequencies across clips using frame per clip (FPC) which denotes the percentage of frames in a clip with the tool present. \DIFadd{We also test our method on YouTube-objects to demonstrate the generalizability and robustness of our proposed method across different types of videos.YouTube-objects dataset is a well-known benchmark in the computer vision community which contains 106 videos for training, and 49 videos for testing, where ten categories ($N=10$) of objects are covered. Each video is split into short-duration clips, and in each clip a few frames are annotated with a bounding box for evaluation.} YouTube-Objects dataset has nearly 100\% FPC for each clip and thus represents the second case (Type II -- WSVOS-COP) demonstrated in Figure \ref{fig_task}. Most categories in Cholec80 based dataset appear on average in less than 60\% of the frames of a particular video labeled with that category, which makes it a much more challenging dataset that represents a \gls{wsvostop} (Type III) problem.
 
\subsection{Experimental Setup and Evaluation Metrics}
   We refrained from data augmentation and present the results trained only with original images in this paper.
 Video frames were resized to 256$\times$256 pixels. Adam optimizer with a learning rate of $1\times10^{-4}$ and a weight decay of $1\times10^{-4}$ was used for training. For the initial nine epochs, only the teacher module and backbone were optimized. Subsequently, the student module was activated alongside the teacher. The $k_1$ value is calculated as 10\% of the number of patches. We fine-tune the $k_2$ value for the ranked-pooling based on data, and the final results reported in this paper are under $k_2=67\%L_v$ ($L_v$ is video length) for the YouTube-Objects data, and $k_2=40\%L_v$ for the surgical data.
We assessed the classification accuracy on frame and video level. For all the models, final segmentations were refined using dense CRF~\cite{krahenbuhl2011efficient}.For the surgical dataset, segmentation performance was evaluated using Intersection over Union (IoU), Dice score and Hausdorff distance (HD). For localization performance we use 
CorLoc (IoU $>$ 50\%) rate \cite{belharbi2023tcam}.


\subsection{Surgical video tool segmentation performance}
\begin{table}[t!]
\caption{\small Comparison to state-of-the-art weakly supervised methods with either frame-level supervision or video-level supervision. Video (V) and Frame (F) accuracy (AC).}
\label{tab_SOTA_compare}
\centering
\setlength\arrayrulewidth{0.9pt}
\setlength\doublerulesep{0.8pt} 
\resizebox{1.0\linewidth}{!}{%
\begin{tabular}{c|c|c|c|c}
\hline
Method & V-AC[\%] $\uparrow$ & F-AC[\%] $\uparrow$& IoU[\%]$\uparrow$ & HD[pixel]$\downarrow$ \\ \hline
MCTformer \cite{xu2022multi} & \textbf{98.27} & 97.63 & 43.14 & 50.07 \\
LayerCAM\cite{jiang2021layercam} & 95.67 & 94.87 & 31.97 & 75.59 \\
XGradCAM\cite{fu2020axiom} & 92.21 & 90.21 & 32.13 & 53.53 \\ \hline
TCAM \cite{belharbi2023tcam} & 97.08 & 95.10 & 28.22 & 54.60 \\
T only (ours) & 98.26 & 97.56 & 47.50 & 48.72 \\
T-S fusion (ours) & 97.11 & 97.23 & 48.64 & 39.61 \\
VDST-Net (ours) & 96.82 & \textbf{98.23} & \textbf{61.80} & \textbf{28.10} \\ \hline
\end{tabular}
}
\end{table}

\begin{figure}[!t]
\includegraphics[width=1.0\linewidth]
{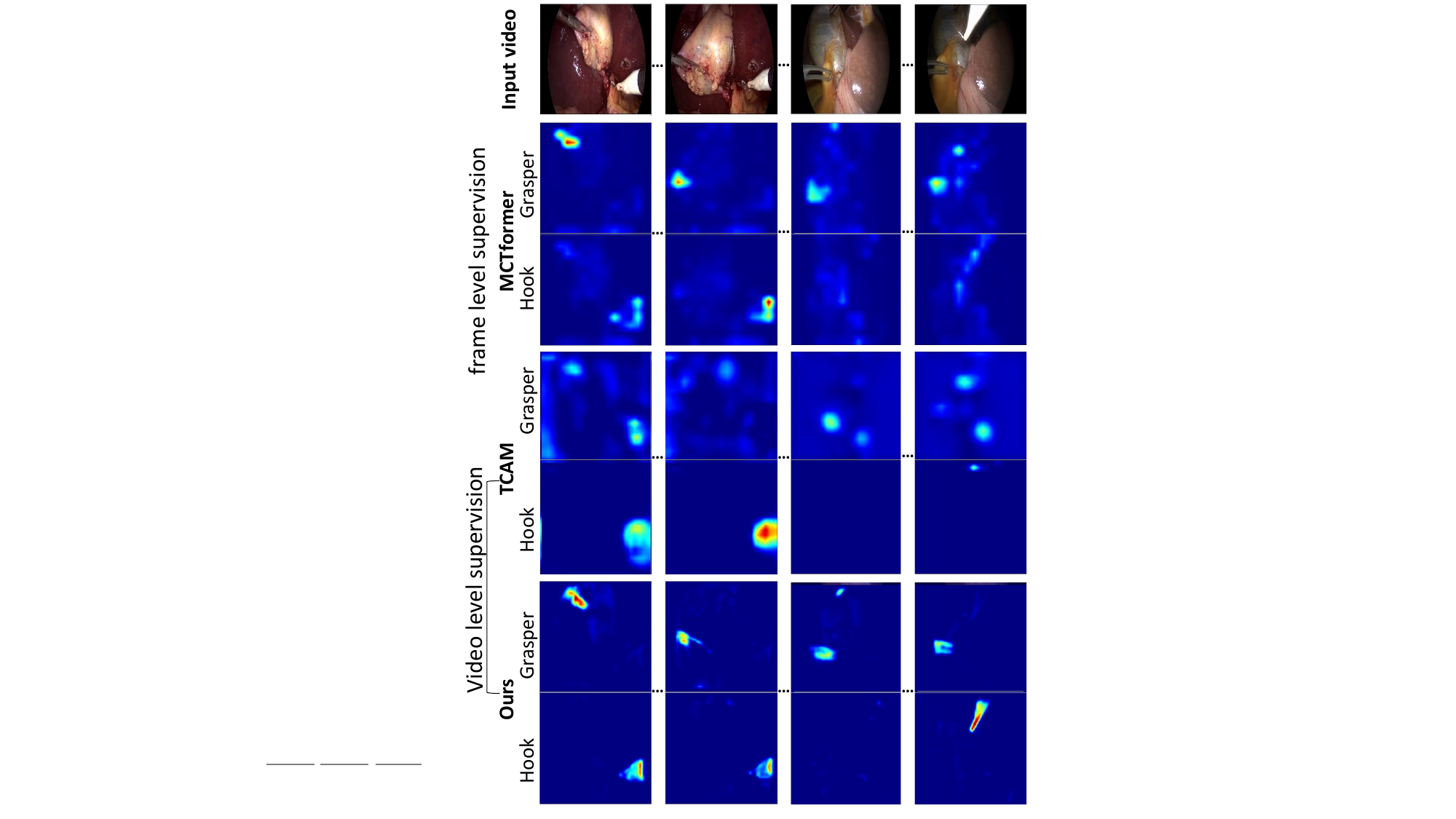}
\centering
\caption{Activation maps of surgical video clips from different methods (MCTformer, TCAM, and our VDST-Net). } 
\label{fig_1}
\end{figure}
On surgical videos, we compare our method to frame-level weakly supervised methods MCTformer \cite{xu2022multi}, LayerCAM \cite{jiang2021layercam}, and XGradCAM \cite{fu2020axiom}, all trained with frame-level labels. Models were adapted to our dataset using their open-source code implementations. We also compare our method to the video-level supervised method TCAM \cite{belharbi2023tcam}, where both our method and TCAM are trained with more challenging video-level class labels. We test our model using teacher (T) only, teacher-student (T-S) fusion, and full~\gls{vstd} variations. For the surgical video data we utilize \DIFdel{another} pre-trained \gls{mae} ViT backbone \cite{he2022masked} which is tuned on a large scale data for segmentation\cite{kirillov2023segment} before it is adapted to the new domain (Cholec80) data.

As presented in Table \ref{tab_SOTA_compare}, MCTformer shows the best performance among frame-level supervision methods. Our teacher-only method, also using a transformer encoder but trained on video-level labels, achieves comparable performance thanks to MLP and our spatio-temporal pooling. Full \gls{vstd} significantly enhances the segmentation performance (61.80\% IoU), with a slight trade-off in video classification score due to potential conflicts between KD loss and classification loss.
 Figure \ref{fig_1} presents activation maps predicted from different methods, \gls{vstd} produces activation maps capable of discerning object shapes more effectively than MCTformer. 

TCAM achieved commendable video and frame accuracy of 97.08\% and 95.10\%, respectively. However, in contrast to detection results on the YouTube-Objects data, its segmentation performance drops compared to our method, indicating a challenge in learning representations for \gls{wsvostop} problems (i.e. in surgical videos).

\subsection{ Analysis on Youtube video object detection Perfomance}

\begin{table*}[th!]
\caption{\small Detection performance on YouTube-Objects \cite{prest2012learning} dataset under CorLoc Metrics. S=Student, T=Teacher
}
\label{tab_ytobj}
\centering
\setlength\arrayrulewidth{0.6pt}
\setlength\doublerulesep{0.8pt} 
\resizebox{1.0\textwidth}{!}{%
\begin{tabular}{cc|cccccccccc|c}
\hline
\multicolumn{2}{c|}{Method}                                                                     & Aero          & Bird          & Boat          & Car           & Cat           & Cow           & Dog           & Horse         & Mbike         & Train         & avg           \\ \hline

\multicolumn{2}{c|}{(Joulin et al., 2014)\cite{joulin2014efficient}}                                          & 25.1          & 31.2          & 27.8          & 38.5          & 41.2          & 28.4          & 33.9          & 35.6          & 23.1          & 25.0          & 31.0          \\
\multicolumn{2}{c|}{(Kwak et al., 2015) \cite{kwak2015unsupervised}}                                            & 56.5          & 66.4          & 58.0          & 76.8          & 39.9          & 69.3          & 50.4          & 56.3          & 53.0          & 31.0          & 55.8          \\
\multicolumn{2}{c|}{(Rochan et al., 2016)\cite{rochan2016weakly}}                                           & 60.8          & 54.6          & 34.7          & 57.4          & 19.2          & 42.1          & 35.8          & 30.4          & 11.7          & 11.4          & 35.8          \\
\multicolumn{2}{c|}{(Tokmakov et al., 2016)\cite{tokmakov2016weakly}}                                        & 71.5          & 74.0          & 44.8          & 72.3          & 52.0          & 46.4          & 71.9          & 54.6          & 45.9          & 32.1          & 56.6          \\
\multicolumn{2}{c|}{(Koh et al., 2016) \cite{koh2016pod}}                                         & 64.3          & 63.2          & 73.3          & 68.9          & 44.4          & 62.5          & 71.4          & 52.3          & 78.6          & 23.1          & 60.2          \\
\multicolumn{2}{c|}{(Tsai et al., 2016)\cite{tsai2016semantic}}                                            & 66.1          & 59.8          & 63.1          & 72.5          & 54.0          & 64.9          & 66.2          & 50.6          & 39.3          & 42.5          & 57.9          \\
\multicolumn{2}{c|}{(Haller and Leordeanu, 2017)\cite{haller2017unsupervised}}                                   & 76.3          & 71.4          & 65.0          & 58.9          & 68.0          & 55.9          & 70.6          & 33.3          & 69.7          & 42.4          & 61.2          \\
\multicolumn{2}{c|}{(Croitoru et al., 2019) \cite{croitoru2019unsupervised}}                  & 81.7          & 51.5          & 54.1          & 62.5          & 49.7          & 68.8          & 55.9          & 50.4          & {33.3} & \textbf{57.0} & 56.5          \\
\multicolumn{2}{c|}{XGradCAM(Fu et al., 2020)\cite{fu2020axiom}}                                 & 68.2 & 44.5 & 45.8 & 64.0 & 46.8 & 86.4 & 44.0 & 57.0 & 44.9 & 45.0 & 54.6       \\
\multicolumn{2}{c|}{LayerCAM(Jiang et al., 2021)\cite{jiang2021layercam}}                                           & 80.0 & 84.5 & 47.2 & 73.5 & 55.3 & 83.6 & 71.3 & 60.8 & 55.7 & 48.1 & 66.0        \\
\multicolumn{2}{c|}{TCAM (Belharbi et al., 2023) \cite{belharbi2023tcam}}                                                                       & 79.4          & {94.9} & {75.7} & 61.7          & {68.8} & \textbf{87.1} & \textbf{75.0} & 62.4          & 72.1          & 45.0          & 72.2          \\ \hline
\multicolumn{1}{c|}{\multirow{3}{*}{VDST-Net + Res34   frozen}}                & T only        & {70.6} & \textbf{97.8} & 44.7          & 78.0          & 31.3          & 57.7          & 53.8          & {42.5} & \textbf{73.9} & 15.6          & 56.6          \\
\multicolumn{1}{c|}{}                                                          & T-S fusion     & 81.6          & 97.7          & 74.4          & 73.9          & 66.7          & 71.4          & 71.8          & 51.6          & 59.5          & 25.0          & 67.4          \\
\multicolumn{1}{c|}{}                                                          & full          & 91.5          & 93.5          & {67.4} & \textbf{82.5} & 63.4          & 80.6          & 64.1          & 56.3          & 55.3          & 36.2          & 69.1          \\ \hline
\multicolumn{1}{c|}{\multirow{3}{*}{VDST-Net + Res34   refined by S}}          & T only        & 88.0          & 95.7          & 74.6          & 69.7          & 78.5          & 80.7          & 57.1          & 62.4          & 55.8          & 20.9          & 68.4          \\
\multicolumn{1}{c|}{}                                                          & T-S fusion      & 87.7          & 95.1          & 76.1          & 70.5          & 81.6          & 81.2          & 58.7          & 61.3          & 58.0          & 22.9          & 69.3          \\
\multicolumn{1}{c|}{}                                                          & full          & 86.4          & 95.1          &\textbf{77.8}        & 71.8          & 76.2          & 77.2          & 54.2          & 55.1          & 57.7          & 22.1          & 67.4          \\ \hline
\multicolumn{1}{c|}{\multirow{3}{*}{VDST-Net + Res34   refined by T}} & T only        & 86.1          & 94.3          & 75.6          & 72.9          & 76.7          & 76.2          & 57.9          & 60.4          & 61.6          & 23.1          & 68.5          \\
\multicolumn{1}{c|}{}                                                          & T-S fusion      & 90.0          & 93.3          & 77.2          & {71.8} & 84.8          & 80.7          & 60.9          & 68.8          & 58.6          & 36.4          & 72.2          \\
\multicolumn{1}{c|}{}                                                          &  {full} & 85.4          & 90.9          & 73.2          & 69.0          & 82.5          & 75.2          & {55.8} & 73.8          & 51.5          & {48.4} & 70.6          \\ \hline
\multicolumn{1}{c|}{\multirow{3}{*}{\textbf{VDST-Net + DINO ViT   (refined by T)}}}         & T only        & 83.8          & 88.6          & 69.5          & 59.3          & 80.7          & 76.6          & 43.2          & 64.9          & 48.7          & 17.8          & 63.3          \\
\multicolumn{1}{c|}{}                                                          & T-S fusion      & 91.0          & 88.2          & 72.7          & 68.7          & \textbf{85.9} & 79.2          & 52.4          & 70.4          & 55.9          & 37.0          & 70.1          \\
\multicolumn{1}{c|}{}                                                          & \textbf{full} & \textbf{92.3} & 91.8          & 72.3          & 70.1          & 82.2          & 82.5          & 73.6          & \textbf{72.6} & 67.8          & 50.7          & \textbf{75.6} \\ \hline
\end{tabular}}
\end{table*}

\begin{figure}[!t]
\includegraphics[width=0.9\linewidth]
{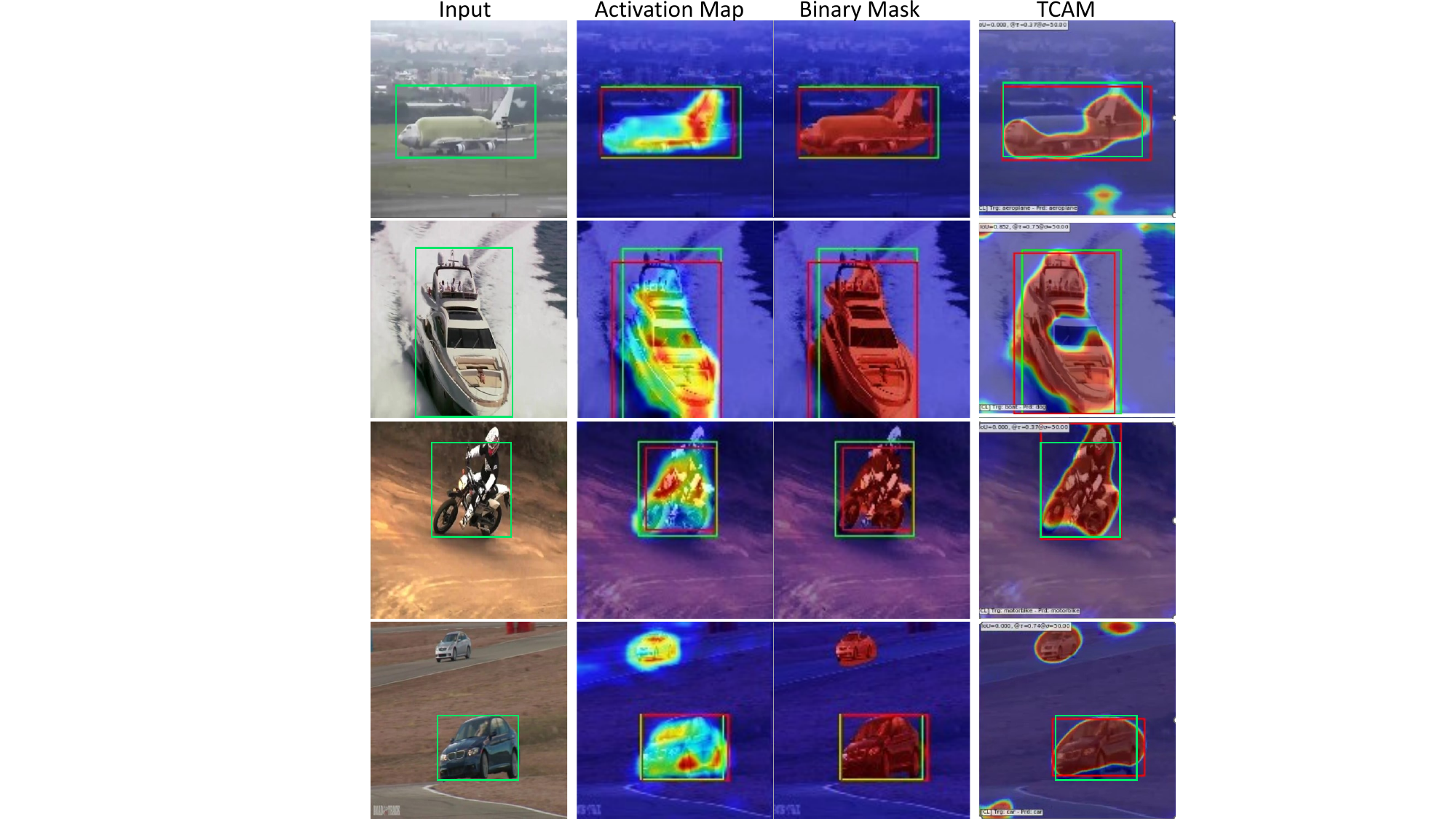}
\centering
\caption{Segmentation and detection performance on Youtube Objects data. The second and third columns are the activation maps and post-processed binary masks of our method. In the last column results are taken from \cite{belharbi2023tcam}.} 
\label{fig_ytobj}
\end{figure}

 As shown in Table \ref{tab_ytobj}, among state-of-the-art methods, TCAM \cite{belharbi2023tcam} leads in performance on the YouTube-Objects dataset for localization. However, qualitatively, it tends to underfit the object contours and over-activate on background, resulting in lower segmentation quality compared to our method, as illustrated in Figure \ref{fig_ytobj}. Our method achieves an overall accuracy improvement from 72.2\% to 75.6\%. Segmentation scores are not reported due to the lack of ground truth masks in the official dataset.

\DIFadd{We utilize the pre-trained DINO ViT}\cite{caron2021emerging} as backbone and fine-tune it on domain-specific data. We also measure performance using a ResNet34 backbone by converting features to \DIFadd{the form of} ViT. We investigate different optimization strategies for the backbone on new domain data: back-propagation through the teacher (denoted as ``refined by T"), through the student (``refined by S"), or keeping the backbone frozen.
Detection outputs from either Teacher (T) or final student (full) or extracted from fused CAM (T-S fusion) are presented. 
Our experiments show that even a frozen ResNet captures a certain level of generalized features. Through our semi-decoupled KD approach, we \DIFadd{improve performance from} 56.6\% to 69.1\%. Interestingly, refining the backbone through the student tends to degrade the final output compared to MLP teacher (68.4\% to 67.4\%), likely due to the challenge of learning generalized features from video-level labels. Optimizing the backbone through the teacher proves to be the effective approach \DIFadd{for both ResNet and ViT backbones leveraging well-learned features}.
For ResNet, we observe that the T-S fusion has a better localization score than the final student (72.2\% vs. 70.6\%), possibly because ResNet features are less generalized and lean more toward the MLP from the teacher module, while the activations learned by the student complement those of the teacher.
In conclusion, we recommend using ViT within our framework and advise against direct student-driven backbone refinements.

\subsection{Ablation Study}

\begin{table*}[t!]
\caption{\small  Ablation study with modifications of overall architecture, spatio-temporal pooling and training strategies. W/O denotes a specific module is removed.}
\label{tab_ablation}
\centering
\setlength\arrayrulewidth{0.9pt}
\setlength\doublerulesep{0.8pt} 
\resizebox{0.7\textwidth}{!}{%
\begin{tabular}{c|c|c|c|c|c|c}
\hline
 & Conditions & Video AC[\%]$\uparrow$ & Frame AC[\%]$\uparrow$ &{IoU[\%]}$\uparrow$& Dice[\%]$\uparrow$ & HD[pixel]$\downarrow$ \\ \hline
\multicolumn{1}{l|}{} & \textbf{VDST-Net full} & 96.82& \textbf{98.23} & \textbf{61.80} & \textbf{67.80}&  \textbf{28.10}\\ \hline
\multirow{2}{*}{Structural ablation} & Teacher only& \textbf{98.41} & 97.56& 47.50& 53.07& 48.72 \\
& Student only& 94.37& 94.76& 42.83& 47.78& 69.45 \\ \hline
\multirow{3}{*}{Training }& Avg pooling& 94.37& 94.51& 53.16& 56.31& 41.65 \\
& Max strategy& 94.95& 95.19& 52.47& 56.09& 40.05 \\ 
& W/O TPC kernel& 96.54& 97.10& 60.46& 63.64& 40.75 \\ \hline
\end{tabular}
}
\end{table*}
\begin{figure*}[t!]
\centering
\includegraphics[width=0.8\textwidth]{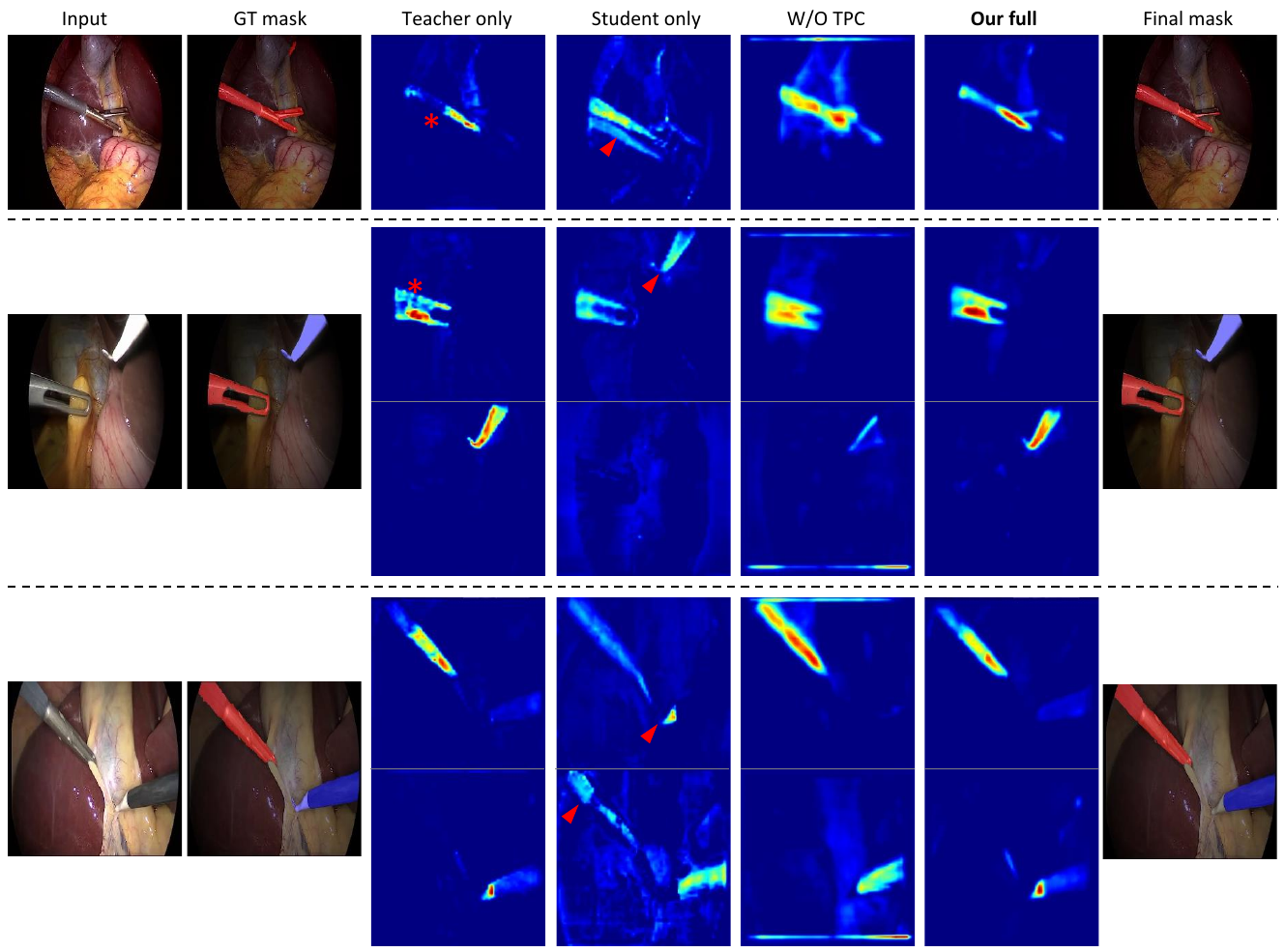}
\caption{ Qualitative results of ablation study and final segmentation masks.} 
\label{fig_2}
\end{figure*}

We evaluated the effects of architecture modifications, pooling variations, and training techniques on model performance (Table \ref{tab_ablation}). Figure \ref{fig_2} shows activation maps from different conditions. Additionally, we demonstrate a further refinement of segmentation generated by using activation maps as prompts for the \gls{sam}~\cite{kirillov2023segment}, showcasing a way of generating semantic pseudo-labels. \\

\noindent{\bf Importance of knowledge distillation:}~~ 
The results highlight the significance of \gls{kd} through teacher and student modules for enhanced segmentation accuracy. A standalone student module, lacking frame-level supervision, shows a notable decline in segmentation performance (Dice score from 67.80\% to 47.64\%), because it predicts error activation by taking information from wrong frames or features, pointed out by red arrowheads in Figure \ref{fig_2}. The teacher module has no such issue but lacks the ability to generate activation maps with good connectivity, as indicated by red asterisks in the figure.\\

\noindent{\bf Effect of ranked top-$k$ pooling:}~~
We \DIFadd{compare} the \DIFadd{ranked top-$k$ pooling to global average and max pooling.} Specifically, with average pooling, we noted a decrease in Dice score by 11.49\%, and with max pooling, the decrease was 11.71\%. Furthermore, there was a considerable decrease in frame classification accuracy, amounting to 3.72\% with average pooling and 3.04\% with max pooling.\\

\noindent{\bf TPC decoupling strategy:}~~
 Additionally, the proposed TPC slice kernel \DIFadd{for activations gating,} enhances the student's learning. The absence of this component leads to a decline in segmentation performance by 4.16\% in Dice and 1.34\% in IoU, as demonstrated in Table \ref{tab_ablation}, due to over-introduced loss from non-target class
knowledge~\cite{zhao2022decoupled}.

\section{Conclusion}
\label{sec:conclusion}

We presented a novel framework named \gls{vstd} for \gls{wsvos} in surgical videos, which is end-to-end trained with coarse video labels. This method uses a teacher-student architecture and \gls{kd} to disentangle spatial and temporal knowledge, minimizing temporal interference for the teacher while enhancing the student's learning capacity.
Experiments on both \gls{cop} and \gls{top} datasets demonstrate our method's efficacy in generating accurate segmentation maps, outperforming state-of-the-art approaches. 

Future work could involve using pseudo masks generated by the student network to train a subsequent lightweight student network (e.g., U-net) for real-time inference. Additionally, our method could be applied to other video datasets beyond the surgical domain, and exploring spatial knowledge captured by other backbones or foundation models for \gls{wsvos} using our method would be interesting.
 We also suggest that, although the student should not directly update the backbone in a shared backbone design, employing separate backbones combined with continual learning strategies such as Exponential Moving Average (EMA) \cite{tobin2017domain} or Elastic Weight Consolidation (EWC) \cite{kirkpatrick2017overcoming} for updating weights could be beneficial.

\section{Acknowledgements.} 
This work was supported by the University of Pennsylvania Thomas B. McCabe and Jeannette E. Laws McCabe Fellow Award and the Linda Pechenik Montague Investigator Award.



{\small
\bibliographystyle{ieee_fullname}
\bibliography{bibliography/refs.bib}
}

\clearpage
\pagenumbering{arabic}

 
\renewcommand*{\thepage}{S\arabic{page}}
 \renewcommand\thefigure{S\arabic{figure}}  
 \renewcommand{\thetable}{S\arabic{table}}
\setcounter{figure}{0} 
\setcounter{table}{0} 

\setcounter{section}{0}
\renewcommand\thesection{\Alph{section}}
\renewcommand\thesubsection{\Alph{section}.\arabic{subsection}}
\newpage
\section{Implementation Details}

\subsection{Teacher student module details}
 
The teacher and student networks have four layers each. The key difference is that each layer of the teacher network uses an MLP, while the student network uses 3D-CNNs for spatio-temporal convolution. A unique feature of the teacher network is that, to help the backbone learn generalized features at different scales, features from each MLP layer are randomly downsampled during training with a probability of 0.5 before the ranked spatial and temporal pooling. This could be considered a heuristic learning strategy that is close to approaches of randomly masking out feature patches [\textcolor{green}{12},\textcolor{green}{26}].  More details about the teacher network's forward operation are shown in the pseudo code in Table \ref{tab_code}. As the student uses 3D-CNN instead of MLP, while keeping the overall architecture the same, student’s network size and computational load are higher than the teacher's, as presented in Table \ref{tab_para}. {link to the code: \url{https://github.com/PCASOlab/VDST-net}}.

\begin{table}[h]
\caption{Numpy-like Pseudo-code for teacher branch forward.}
\label{tab_code}
\centering
\begin{tabular}{|l|}
\hline
\begin{lstlisting}[
    language=Python, 
    basicstyle=\ttfamily\scriptsize,  % Change \footnotesize to your desired size
    commentstyle=\color{cyan}\scriptsize % Ensure comments match the font size
]
# image_encoder - ResNet or Vision Transformer
# W_MLP [c_in, c_e]- one layer of proj of the MLP  
# W_fc [c_e, N] - learned proj of feature to CAM/Class
# v[bz, d, h, w ,c] - minibatch of videos
# labels[bz, N] - minibatch of ground truth class

# extract feature representations of video 
T_f = image_encoder(v) #[bz, d, h_f, w_f, c_f]
T_e = T_f.reshape[bz* d* h_f* w_f, c_f]

for W_MLP in MLP_layers:
    # One layer of MLP 
    T_e = ReLU(np.dot(T_e, W_MLP), axis=1)
    #  prob 0.5 of downsample
    if Random(0,1) > 0.5:
       T_e.reshape[bz, d, h_new, w_new, c_e] 
       # avg pool with kernel k, and stride s
       T_e = Avg_pool_3D (T_e,k=[1,2,2],s=[1,2,2]) 
       T_e.reshape[bz* d* h_new* w_new* c_e] 

T_e.reshape[bz, d, h_new, w_new, c_e] 
# intepolate in spatial 
T_e = 3D_interpolate (T_e) #[bz, d, h_f, w_f, c_e]
#ranked spatial pooling
slice_valid = Rank_pool_hw(T_e,k1) # [bz,d,1,1,c_e]
#ranked temporal pooling
final_f = Rank_pool_d(slice_valid,k2) # [bz,1,1,1,c_e]
final_f.reshape(bz,c_e)

#loss for training
class_logit = sigmoid (np.dot(final_f, W_fc),axis=1)
loss_t = BCE_loss(class_logit, labels, axis=1)
#ST_CAM interface, N class activation maps
T_e.reshape[bz* d* h* w, c_e] 
ST_CAM = RelU(np.dot(T_e, W_fc),axis=1)  
ST_CAM.reshape[bz, d, h_f, w_f, N]
\end{lstlisting} \\
\hline
\end{tabular}
\end{table}

\begin{table}[t!]
\caption{\small Teacher and student parameter overview with video feature input. FLOPs: Floating-point operations per second. }
\label{tab_para}
\centering
\setlength\arrayrulewidth{0.9pt}
\setlength\doublerulesep{0.8pt} 
\resizebox{0.9\linewidth}{!}{%
\begin{tabular}{ccc}
\hline
Parameter                  & teacher    & student   \\
Size                       & 4.02 MB    & 17.02 MB  \\
Trainable params           & 1.05 M     & 4.46 M    \\
Forward/backward pass size & 1325.39 MB & 2092.8 MB \\
FLOPs                      & 31.48 G    & 132.68 G  \\\hline
\end{tabular}
}
\end{table}

\subsection{Training details}

We implemented our training on an Nvidia A5500-based compute platform. Both the teacher and student networks contain a dropout layer after either the MLP (in the teacher) or the 3D-CNN (in the student). For the first nine epochs, only the teacher network was optimized with a dropout rate of 0.5. Subsequently, the student module was enabled with a dropout rate of 0.5, while the teacher's dropout was disabled. After this, both networks were trained for 30 epochs. Optionally, either the teacher or the student can optimize the backbone. For instance, if the teacher is allowed to fine-tune the backbone, the gradient flow from the student to the backbone is stopped.

\section{Transient Object Presence Cholec80 Data Statistics }
\begin{table}[t!]
\caption{\small  Statistics of the Transient Object Presence (TOP) Cholec80 Data.  CholecSeg8K removal: Removal of overlapping clips annotated by CholecSeg8K from the original Cholec80 dataset, as described in section 4.1. }
\label{tab_data}
\centering
\setlength\arrayrulewidth{0.6pt}
\setlength\doublerulesep{0.8pt} 
\resizebox{0.9\linewidth}{!}{%
\begin{tabular}{l|rr|rrr}
\hline
             & \multicolumn{2}{c|}{Cholec80} & \multicolumn{3}{c}{CholecSeg8K removal} \\ \cline{2-6} 
Category     & Frames        & Clips      & Frames      & Clips     & FPV         \\ \hline
Grasper      & 102k          & 5k         & 83k         & 4k        & 63.6\%      \\
Bipolar      & 9k            & 0.59k      & 7k          & 0.48k     & 48.2\%      \\
Hook         & 103k          & 4k         & 87k         & 3k        & 80.8\%      \\
Scissor      & 3k            & 0.24k      & 3k          & 0.19k     & 46.6\%      \\
Clipper      & 6k            & 0.35k      & 5k          & 0.28k     & 57.0\%      \\
Irrigator    & 10k           & 0.62k      & 8k          & 0.51k     & 52.7\%      \\
Specimen bag & 11k           & 0.62k      & 9k          & 0.51k     & 60.7\%      \\ \hline
\end{tabular}}
\end{table}

Details of the complete Cholec80 video clips set and the sampled training set are presented in Table \ref{tab_data}.  Certain instruments, like the Hook, have a high presence frequency  throughout 
 the overall dataset or within video. The latter is quantified by the percentage of frames per video (FPV). Instruments such as Scissor and Bipolar appear less frequently, with FPV percentages around or below 60\%.
\section{Additional Results}

\begin{figure*}[t!]
\includegraphics[width=1.0\textwidth]{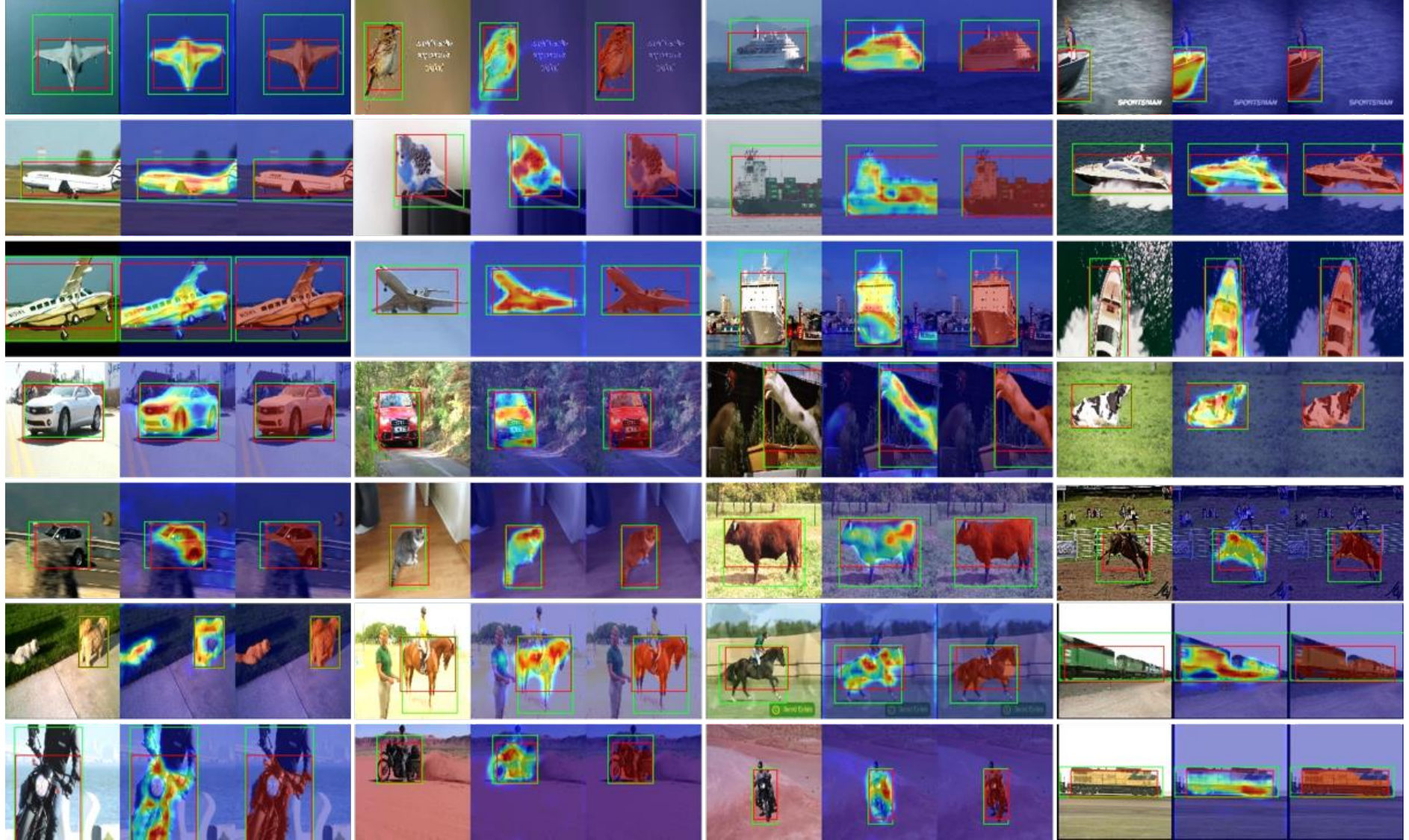}
\centering
\caption{More qualitative results of the proposed method on Youtube-objects data. Original images, activation maps, and masks after thresholding are presented sequentially. Our method demonstrates robustness, producing activation maps and segmentation masks that accurately follow the objects' contours, with minimal false activation on the background.  
} 
\label{fig_s}
\end{figure*}
\begin{table}[h!]
\caption{\small Results of using separate backbones for teacher and student respectively,  CorLoc score on the Youtube-Objects data are reported.}
\label{tab_mix}
\centering
\setlength\arrayrulewidth{0.9pt}
\setlength\doublerulesep{0.8pt} 
\resizebox{0.7\linewidth}{!}{%
\begin{tabular}{llc}
\hline
                          & Full          & T only\\
Res teacher + ViT student & \textbf{75.0}   &68.5         \\
ViT teacher + Res student & 68.4           & 63.3       \\ \hline
\end{tabular}
}
\end{table}
\begin{table}[t!]
\caption{\small  Results of using Resnet34 as feature extractor on Cholec80 data.}
\label{Tab_resnet}
\centering
\setlength\arrayrulewidth{0.9pt}
\setlength\doublerulesep{0.8pt} 
\resizebox{0.7\linewidth}{!}{%
\begin{tabular}{lccc}
\hline
         & IoU{[}\%{]} $\uparrow$ & Dice{[}\%{]$\uparrow$} & HD{[}pix{]$\downarrow$} \\
VDST-Net & 43.58       & 46.15        & 53.50       \\ \hline
\end{tabular}
}
\end{table}

We also present results using a ViT backbone for the student and a ResNet backbone for the teacher, as well as the reverse configuration. As shown in Table \ref{tab_mix}, both configurations improve performance over the teacher alone. Notably, using ViT as the student results in the highest performance among them.

Additionally, we utilize a ResNet-34 backbone for the TOP cholec80 videos (Table \ref{Tab_resnet}). While this approach shows a decline in accuracy compared to our ViT-based method, it still achieves comparable performance to other state-of-the-art methods, as presented in Table 2.

More qualitative results on the YouTube-Objects dataset are presented in Figure \ref{fig_s}.

\end{document}